\documentclass[twoside,11pt]{article}

\usepackage{jmlr2e}
\usepackage{algorithm}
\usepackage{algpseudocode}
\usepackage{amsmath}
\usepackage{subcaption}
\usepackage{listings}
\usepackage{xcolor}
\usepackage[toc,page]{appendix}

\ShortHeadings{AdamD: Improved bias-correction in Adam}{St. John}
\firstpageno{1}

\begin{document}

\title{AdamD: Improved bias-correction in Adam}

\author{\name John St John \email john@ravel.bio \\
       \addr Ravel Biotechnology \\
       2325 3rd St, Suite 328 \\
       San Francisco, CA 94017, USA
}

\editor{NA}

\maketitle

\begin{abstract}
Here I present a small update to the bias-correction term in the Adam optimizer that has the advantage of making smaller gradient updates in the first several steps of training. With the default bias-correction, Adam may actually make larger than requested gradient updates early in training. By only including the well-justified bias-correction of the second moment gradient estimate, $v_t$, and excluding the bias-correction on the first-order estimate, $m_t$, we attain these more desirable gradient update properties in the first series of steps. The default implementation of Adam may be as sensitive as it is to the hyperparameters $\beta_1, \beta_2$ partially due to the originally proposed bias correction procedure, and its behavior in early steps.

\end{abstract}


\section{Description and proposal}

Adam, and other common optimizers dervived from it since \citep{loshchilov:2019}, use a bias-correction factor of $1-\beta_1^t$ and $1-\beta_2^t$. As others have previously noted \citep{kingma:17, mann:19, you:20}, these bias correction terms can equivalently be viewed as a modification to the learning rate of $\frac{\sqrt{1-\beta_2^t}}{1-\beta_1^t}$ (also see Appendix \ref{adam:equivalence}). The originally stated goal of the bias-correction factor was at least partially to reduce the initial learning rate in early steps, before the moving averages had been well initialized \citep{kingma:17, mann:19}. When thought of as a learning rate warm up period, you can also interrogate whether this learning rate warm up period is well behaved. For example does it strictly increase from a low rate to a high rate as the timestep $t$ increases? In this write-up I will show that the default formulation of Adam's bias-correction factor, with recommended settings for $\beta_1, \beta_2$ is not well behaved in the beginning. This term first decreases for several steps, then later increases, eventually converging at the requested learning rate. With some values of $\left(\beta_1, \beta_2\right)$, I show (figure \ref{fig:1}) that this term can even lead to a higher than requested learning rate in the first set of steps, which goes against one of the stated goals for the bias correction term \citep{loshchilov:2019}.

The main reason for this observed issue is the inclusion of the bias correction term in the numerator, on the first-moment estimate $m_t$. The bias correction terms inflate the variables they are applied to. This is easy to see because $0 < \beta^t < 1$ so $1/\beta^t >= 1$. Applying the warm-up bias correction to the denominator (the estimated variance of the gradient), $v_t$ is well argued in \cite{kingma:17}. In fact little justification for inclusion is provided, other than that estimating the term on $m_t$ is similar and easy to show \citep{kingma:17}. Since the correction applied to $v_t$ is in the denominator, this term by itself have the desired impact of allowing the algorithm to catch up with it's estimates of the un-centered variance, and prevent early step sizes that are too large early in training due to the estimate being initialized to 0 \citep{kingma:17}. Including the bias correction term on the numerator, on the estimate of the mean of the gradients however, will have the opposite (anti-conservative) effect. This can actually lead to larger initial step sizes than the requested learning rate $\alpha$, depending on the values of $\beta_1, \beta_2$. Even at default suggested values of $\beta_1, \beta_2$, the inclusion of this term in the numerator has a non-monotonic effect on the learning rate early in training that seems difficult to justify (see \ref{fig:1}).

Making the small change of only including the bias correction term to the variance, $v_t$ to the Adam optimizer results in a cleaner learning rate increase at a variety of different values of $\beta_1, \beta_2$ (Figure \ref{fig:1}).

\begin{figure}[h] \label{fig:1}
  \begin{subfigure}{0.5\textwidth}
    \includegraphics[width=\linewidth]{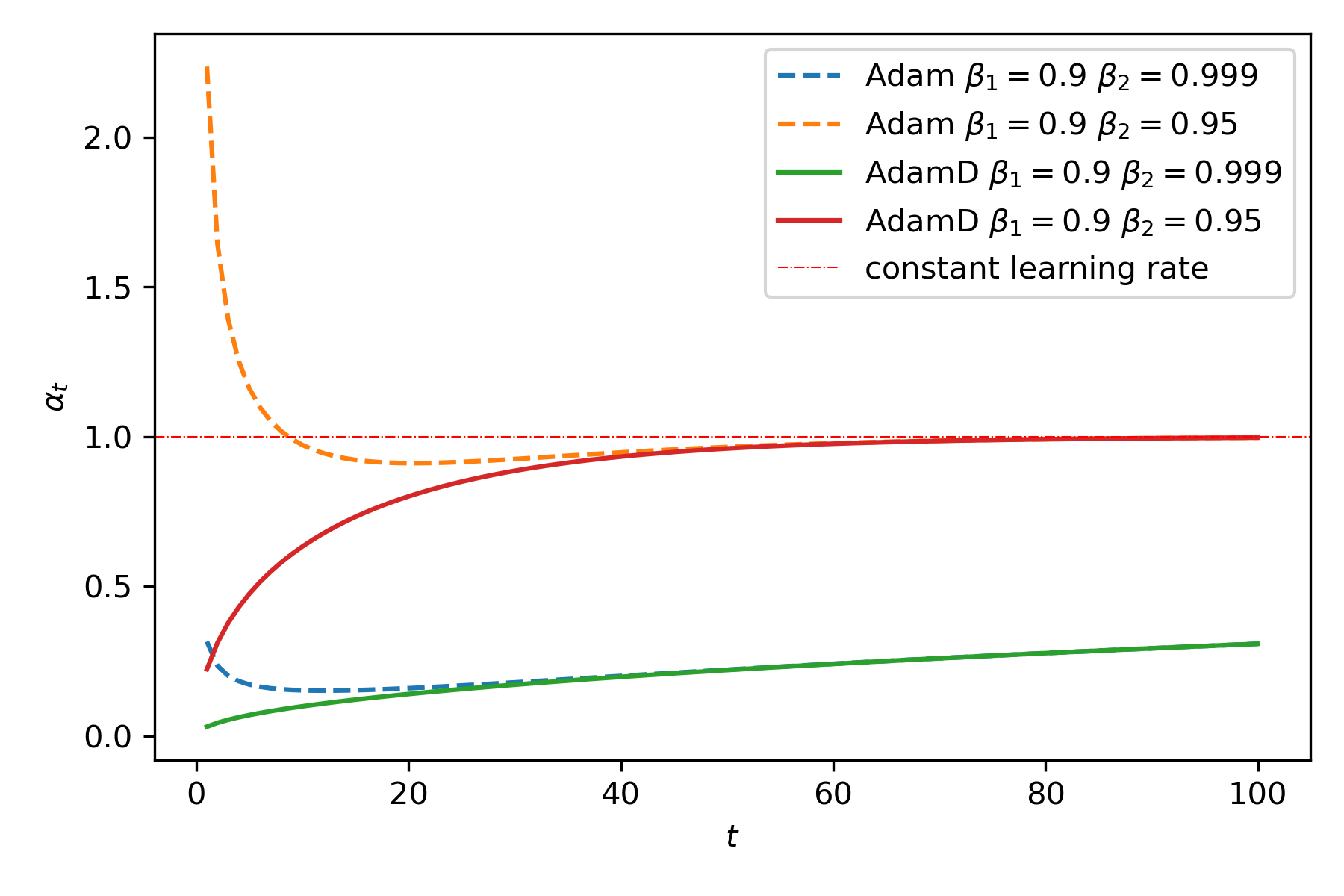}
    \caption{$\alpha_t$ for the first 100 steps} \label{fig:1a}
  \end{subfigure}%
  \hspace*{\fill}   
  \begin{subfigure}{0.5\textwidth}
    \includegraphics[width=\linewidth]{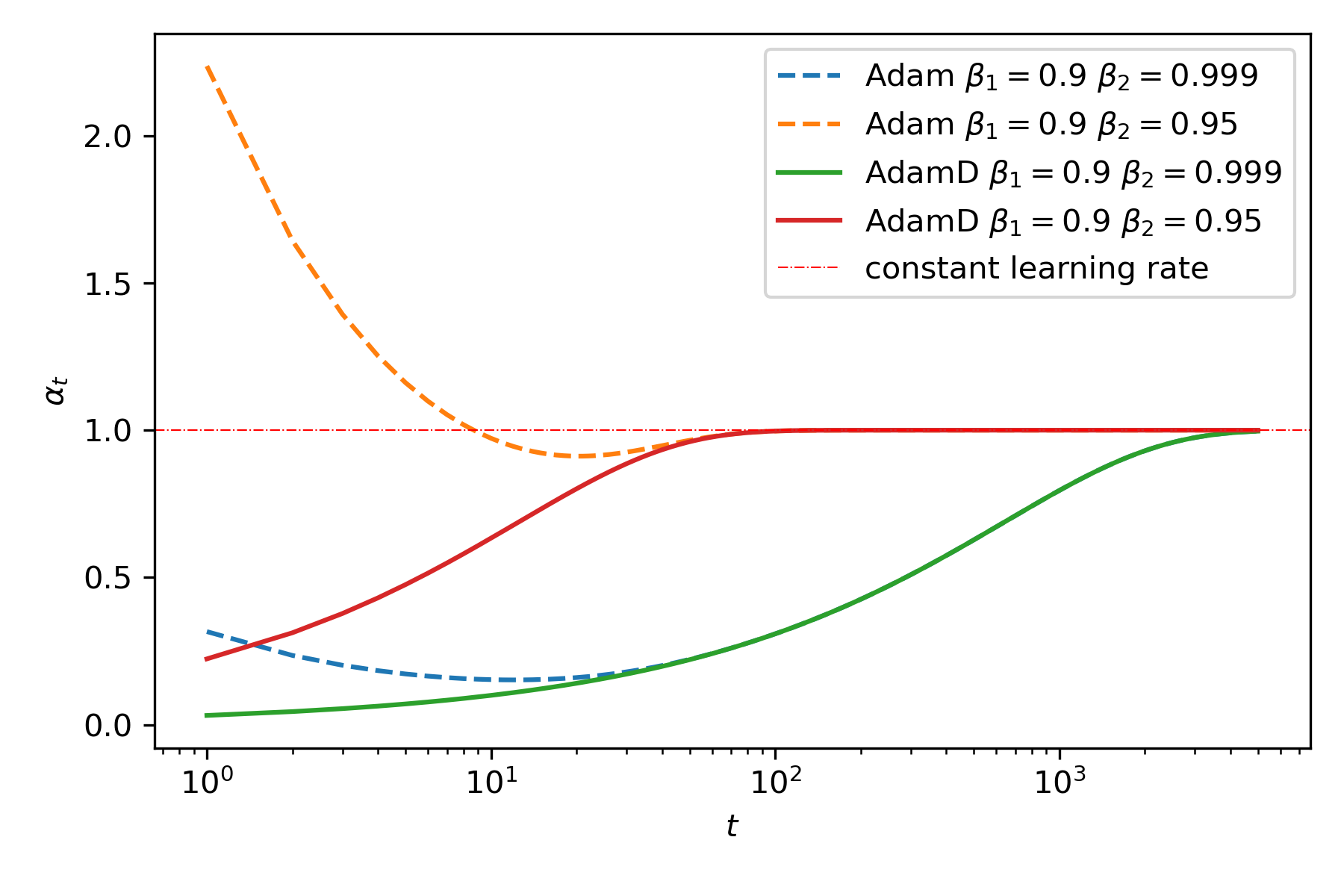}
    \caption{$\alpha_t$ for the first 5000 steps} \label{fig:1b}
  \end{subfigure}%

\caption{As you can see, the $\alpha_t$ rule in Algorithm \ref{AdamD} (AdamD) starts small and increases with time, vs the default $\alpha_t$ in Adam which has an initial period of decrease. Note that at some values of $\beta_1, \beta_2$ this learning rate can be a large increase over the baseline learning rate at the beginning, which defeats the original purpose of this term, and may explain some of the algorithm's sensitivity to choice of $\beta_1, \beta_2$.} \label{fig:1}
\centering
\end{figure}

\begin{algorithm}
\caption{Adam as originally described in \cite{kingma:17} but their proposed alternative (equivalent, see Appendix \ref{adam:equivalence}, or the note right before section 2.1 in \cite{kingma:17}) formula that highlights the learning rate as a function of the timestep.} \label{Adam}
\begin{algorithmic}
\Require $\alpha$ \Comment{Stepsize}

\Require $\beta_1, \beta_2 \in [0, 1)$ \Comment{Exponential decay rates for the moment estimates.}

\Require $f(\theta)$ \Comment{Stochastic objective function with parameters $\theta$}

\Require $\theta_0$ \Comment{Initial parameter vector.}

\State $m_0 \gets 0$ \Comment{Initialize first moment vector}

\State $v_0 \gets 0$ \Comment{Initialize second moment vector} 

\State $t \gets 0$ \Comment{Initialize timestep}

\While{ $\theta_t$ not converged}
\State $t \gets t+1$
\State $g_t \gets \nabla_\theta f_t(\theta_{t-1})$ \Comment{Get gradients w.r.t. stochastic objective at timestep $t$}

\State $m_t \gets \beta_1 \cdot m_{t-1} + (1-\beta_1) \cdot g_t$ \Comment{Update biased first moment estimate}

\State $v_t \gets \beta_2 \cdot v_{t-1} + (1 - \beta_2) \cdot g_t^2$ \Comment{Update biased second raw moment estimate}

\State \textcolor{blue}{$\alpha_t := \alpha \cdot \left(\frac{\sqrt{1-\beta_2^t}}{1-\beta_1^t}\right)$ \Comment{Set the learning rate for step $t$ with the original debiasing term.}}

\State $\theta_t \gets \theta_{t-1} - \alpha_i \cdot \left(\frac{m_t}{\sqrt{v_t} + \epsilon}\right)$ \Comment{Update parameters.}

\EndWhile

\end{algorithmic}
\end{algorithm}

\begin{algorithm}
\caption{AdamD, identical to Algorithm \ref{Adam} other than the highlighted learning rate by timestep modification ($\alpha_t$)} \label{AdamD}
\begin{algorithmic}
\Require $\alpha$ \Comment{Stepsize}

\Require $\beta_1, \beta_2 \in [0, 1)$ \Comment{Exponential decay rates for the moment estimates.}

\Require $f(\theta)$ \Comment{Stochastic objective function with parameters $\theta$}

\Require $\theta_0$ \Comment{Initial parameter vector.}

\State $m_0 \gets 0$ \Comment{Initialize first moment vector}

\State $v_0 \gets 0$ \Comment{Initialize second moment vector} 

\State $t \gets 0$ \Comment{Initialize timestep}

\While{ $\theta_t$ not converged}
\State $t \gets t+1$
\State $g_t \gets \nabla_\theta f_t(\theta_{t-1})$ \Comment{Get gradients w.r.t. stochastic objective at timestep $t$}

\State $m_t \gets \beta_1 \cdot m_{t-1} + (1-\beta_1) \cdot g_t$ \Comment{Update biased first moment estimate}

\State $v_t \gets \beta_2 \cdot v_{t-1} + (1 - \beta_2) \cdot g_t^2$ \Comment{Update biased second raw moment estimate}

\State \textcolor{blue}{$\alpha_t := \alpha \cdot \sqrt{1-\beta_2^t}$ \Comment{Set the learning rate for step $t$ with the new debiasing term.}}

\State $\theta_t \gets \theta_{t-1} - \alpha_t \cdot \left(\frac{m_t}{\sqrt{v_t} + \epsilon}\right)$ \Comment{Update parameters.}

\EndWhile

\end{algorithmic}
\end{algorithm}

\section{Conclusion and next steps}
Here I show a small modification to the Adam algorithm that behaves better in the initial series of steps of training. The proposed modification better accomplishes the goal \citep{mann:19, kingma:17} of starting out with a small learning rate when the model is first initialized, and then quickly ramping up the learning rate to the desired $\alpha$. Additionally this bias correction simplification accomplishes all of the theoretical objectives stated in section 3 of the original Adam publication. As some other authors have pointed out, replacing this bias correction term completely with a learning rate warm-up period is another reasonable option \citep{you:20}. The $\beta$ parameters of Adam should be re-explored with this modified update rule in place to see what changes this new update procedure has in practice.

This modification to Adam may be easily applied to any of the variant algorithms on Adam, such as AdamW \citep{loshchilov:2019}, Lamb \citep{you:20}, and others. Future work should explore what effect, if any, this modification has on the peformance of algorithms that make use of bias correction in the numerator of the gradient update.


\acks{
    I would like to acknowledge my colleagues at Ravel Biotechnology, especially Nathan Boley and Omid Shams Solari, for useful discussion, as well as Ben Mann for his medium post \citep{mann:19} which lead to these observations.
}


\newpage

\bibliography{sample}
\newpage

\begin{appendices}

\section{Learning rate modification formulation of Adam bias-correction.} \label{adam:equivalence}



In this section we show the alternative formulation of the update rule for Adam as a learning rate scheduling term. To show this we temporarily remove the $\epsilon$ term from the denominator, and then add it back in at the end. This observation has been made previously \citep{kingma:17, you:20}, however here we spell it out.

\begin{subequations}
\begin{align}
 \theta_t \gets & \alpha \cdot \hat{m}_t / \left(\sqrt{\hat{v}_t}+\epsilon\right) & \text{Orignal adam update rule}\\
 \approx & \alpha \cdot \hat{m}_t / \left(\sqrt{\hat{v}_t}+\epsilon\right) & \text{Temporarily drop the $\epsilon$}\\
  = & \alpha \cdot \frac{m_t / \left( 1-\beta_1^t \right) }{\sqrt{v_t/\left( 1-\beta_2^t\right)}} & \text{Substitute back in the bias corrections} \\
  = & \alpha \cdot \frac{m_t \cdot \sqrt{1-\beta_2^t}}{\sqrt{v_t}\cdot\left( 1-\beta_1^t \right)} & \text{Simplify the fraction} \\
  = & \alpha \cdot \frac{m_t}{\sqrt{v_t}} \cdot \frac{\sqrt{1-\beta_2^t}}{1-\beta_1^t} \\
  = & \left(\alpha \cdot \frac{\sqrt{1-\beta_2^t}}{1-\beta_1^t}\right) \cdot \frac{m_t}{\sqrt{v_t}} & \text{Show the bias corrections define $\alpha_t$} \\
  \approx & \left(\alpha \cdot \frac{\sqrt{1-\beta_2^t}}{1-\beta_1^t}\right) \cdot \frac{m_t}{\sqrt{v_t} + \epsilon} & \text{Add back the $\epsilon$ term} \label{adam:lr_update_equivalence}
\end{align}
\end{subequations}

\section{Code for figures}
This section includes code used to generate the figures in this publication.

\subsection{Code for figure \ref{fig:1a}.}

\begin{lstlisting}[language=Python]
import math
import numpy as np
from matplotlib import pyplot as plt
import matplotlib

def get_debias_original(step, beta1, beta2):
    bc1 = 1-beta1**step
    bc2 = 1-beta2**step
    return math.sqrt(bc2) / bc1

def get_debias_update(step, beta1, beta2):
    bc2 = 1-beta2**step
    return math.sqrt(bc2)


x=np.arange(100)+1

for db_func, b1, b2, desc, style in [
    (get_debias_original, 0.9, 0.999, "Adam", "--"),
    (get_debias_original, 0.9, 0.95, "Adam", "--"),
    (get_debias_update, 0.9, 0.999, "AdamD", "-"),
    (get_debias_update, 0.9, 0.95, "AdamD", "-"),
]:
    plt.plot(
        x, 
        [db_func(s, beta1=b1, beta2=b2) for s in x], 
        label=rf"{desc} $\beta_1={b1}$ $\beta_2={b2}$", 
        linestyle=style
    )
ax = plt.gca()
ax.axhline(
    1.0, 
    label="constant learning rate", 
    color="red", 
    linestyle="-.", 
    lw=0.5,
)
plt.xlabel(r"$t$")
plt.ylabel(r"$\alpha_t$")
plt.legend()
plt.tight_layout()
plt.savefig("Zoomed_Adam_v_AdamD.png", dpi=300)
\end{lstlisting}

\subsection{Code for figure \ref{fig:1b}.}

\begin{lstlisting}[language=Python]
x=np.arange(5000)+1

for db_func, b1, b2, desc, style in [
    (get_debias_original, 0.9, 0.999, "Adam", "--"),
    (get_debias_original, 0.9, 0.95, "Adam", "--"),
    (get_debias_update, 0.9, 0.999, "AdamD", "-"),
    (get_debias_update, 0.9, 0.95, "AdamD", "-"),
]:
    plt.plot(
        x, 
        [db_func(s, beta1=b1, beta2=b2) for s in x], 
        label=rf"{desc} $\beta_1={b1}$ $\beta_2={b2}$", 
        linestyle=style
    )
    ax = plt.gca()
    ax.set_xscale('log')
ax.axhline(
    1.0, 
    label="constant learning rate", 
    color="red", 
    linestyle="-.", 
    lw=0.5,
)
plt.legend()
plt.xlabel(r"$t$")
plt.ylabel(r"$\alpha_t$")
plt.tight_layout()
plt.savefig("Log_Adam_v_AdamD.png", dpi=300)
\end{lstlisting}

\end{appendices}

\end{document}